\documentclass[letterpaper, 10 pt, conference]{ieeeconf}  

\IEEEoverridecommandlockouts                              

\usepackage{times}
\usepackage{epsfig}
\usepackage{graphicx}
\usepackage{amsmath}
\usepackage{amssymb}
\usepackage{tablefootnote}
\usepackage{multirow}
\usepackage{xcolor} 
\usepackage{verbatim}
\usepackage{url}
\usepackage{bm}

\usepackage{caption}
\usepackage{adjustbox}
\usepackage{balance}
\usepackage{float}

\title{\LARGE \bf
Point Cloud Completion by Learning Shape Priors
}

\author{Xiaogang Wang$^{1}$, Marcelo H Ang Jr$^{1}$ and Gim Hee Lee$^{2}$
\thanks{$^{1}$Department  of  Mechanical  Engineering,  National  University  of  Singapore
        {\tt\small xiaogangw@u.nus.edu, mpeangh@nus.edu.sg}}%
\thanks{$^{2}$Computer Vision and Robotic Perception (CVRP) Lab, Department of Computer Science, National University of Singapore
        {\tt\small dcslgh@nus.edu.sg}}%
}

\begin{document}

\maketitle
\thispagestyle{empty}
\pagestyle{empty}

\begin{abstract}
In view of the difficulty in reconstructing object details in point cloud completion, we propose a shape prior learning method for object completion. The shape priors include geometric information in both complete and the partial point clouds. We design a feature alignment strategy to learn the shape prior from complete points, and a coarse to fine strategy to incorporate partial prior in the fine stage. To learn the complete objects prior, we first train a point cloud auto-encoder to extract the latent embeddings from complete points. Then we learn a mapping to transfer the point features from partial points to that of the complete points by optimizing feature alignment losses. The feature alignment losses consist of a L2 distance and an adversarial loss obtained by Maximum Mean Discrepancy Generative Adversarial Network (MMD-GAN). The L2 distance optimizes the partial features towards the complete ones in the feature space, and MMD-GAN decreases the statistical distance of two point features in a Reproducing Kernel Hilbert Space. We achieve state-of-the-art performances on the point cloud completion task. Our code is available at \url{https://github.com/xiaogangw/point-cloud-completion-shape-prior}.
\end{abstract}

\section{INTRODUCTION}
Real-world 3D data collected by LiDAR are often sparse, incomplete and non-uniform distributed, in which both important geometric and semantic information are missing. Point cloud completion has far-reaching applications on robotics and perception. For example,  localization and mapping are done under the partial and sparse point clouds for the autonomous driving system~\cite{lee2013structureless}, it would be beneficial for the Simultaneous Localization and Mapping (SLAM)~\cite{mur2015orb,engel2014lsd} or Structure-from-Motion (SfM) \cite{agarwal2011building} systems if we can apply the completion technique to obtain the complete point cloud first.
To this end, we propose a point completion network to learn the shape priors to generate complete and dense point clouds.

Many deep Convolutional Neural Networks (CNN)
are proposed to learn semantic meaningful features from images for image retrieval~\cite{lin2015deep} or transformation~\cite{isola2017image}.
Some works~\cite{hu2018cvm,melekhov2016siamese,feng20192d3d} propose Siamese network architectures to match images features between two different modalities. 
Several approaches~\cite{isola2017image,liu2017unsupervised,zhu2017unpaired} adopt generative adversarial network (GAN)~\cite{goodfellow2014generative} to transfer image from one domain to the other. 
Even though extensive works are proposed for image transformation, limited research has been conducted for unorganized 3D point clouds.

\begin{figure}
  \includegraphics[width=\linewidth]{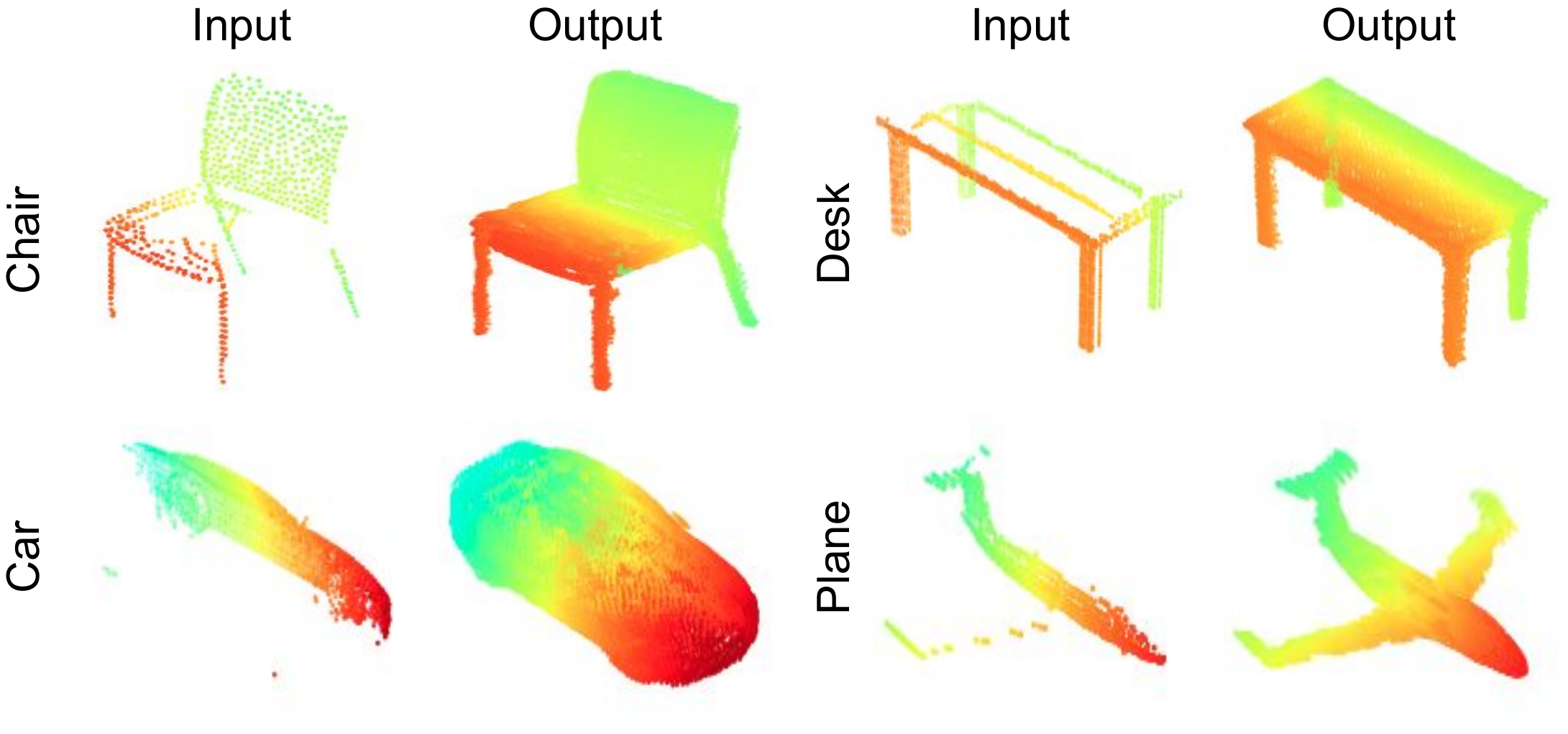}
  \caption{Our point cloud completion network is able to reconstruct complete and dense objects with finer details. The resolutions of input partial points and synthesized output points are 2048 and 16384, respectively.}
  \label{task_picture}
  \vspace{-0.7cm}
\end{figure}
Learning representations to point clouds is very challenging due to the sparseness, incompleteness and unorderness of the 3D points.
The pioneering work PointNet~\cite{qi2017pointnet} addressed the unordered problem by using a shared multi-layer perceptron (MLP) and symmetrical max-pooling operation to achieve the permutation-invariant property, and they showed impressive results on the point clouds classification and segmentation tasks. However, most existing deep networks are designed for tasks that take complete and clean point clouds in synthetic dataset as inputs, which lead to unreliable outputs on most 3D point clouds collected from real world that are usually incomplete and sparse. 
To alleviate this problem, several works are proposed for point cloud completion~\cite{yuan2018pcn,topnet2019,liu2019morphing} task. Although they have achieved impressive results on reconstructing object shapes, the synthesized objects are not realistic enough and lack object details.
To this end, we propose a simple yet effective approach by leveraging the shape priors in the completion process.

In this paper, we propose feature alignment strategies in a coarse-to-fine pipeline to learn shape priors from both complete and incomplete 3D point clouds simultaneously.

More specifically,
we first train an auto-encoder~\cite{yuan2018pcn} on complete point clouds from the training dataset, then we extract one 256-dim and one 1024-dim point features as the latent embeddings of complete objects.
These two features are strong priors from the complete 3D shapes~\cite{mandikal20183d}. 
Our feature alignment strategies include a feature matching loss~\cite{johnson2016perceptual,mandikal20183d} and an adversarial loss, which is able to transfer the incomplete point features to the target complete space. 
Feature matching loss leads the partial point features to the complete ones by optimizing the L2 distance on the latent embeddings.
The adversarial learning with the Maximum Mean Discrepancy (MMD) distance is used to minimize the statistical distribution distance between two features in a Reproducing Kernel Hilbert Space (RKHS). 
In order to preserve the details of partial inputs, we concatenate the input points with the reconstructed coarse output and refine the combined points in the finer stage inspired by~\cite{liu2019morphing,Wang_2020_CVPR}.
Finally, we jointly optimize the Chamfer Distance (CD), feature alignment losses and the adversarial loss in an end-to-end manner to synthesize the dense and complete point clouds.
Fig.~\ref{task_picture} shows examples of our method on the completion task. 

\noindent Our contributions are as follows:
\begin{itemize}
    \item We propose a point completion network to learn the object shape priors to reconstruct the complete point clouds.
    \item We propose 3D feature alignment methods to learn the shape priors from both the complete and incomplete point clouds.
    \item We achieve superior results on both synthetic and real-world datasets compared to existing approaches.
\end{itemize}

\begin{figure*}
\centering
  \includegraphics[width=1\linewidth]{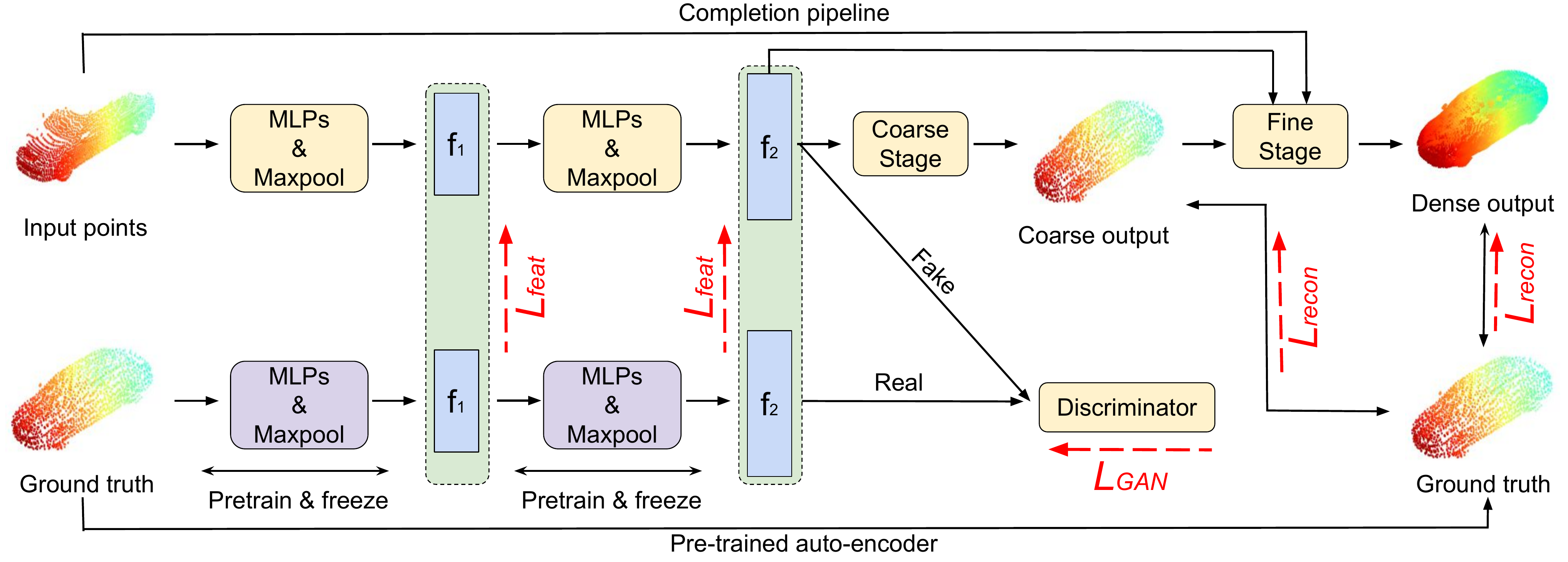} 
  \vspace{-6mm}
  \caption{
  An illustration of the overall pipeline. We first train a point cloud auto-encoder (lower part) and fix the weights to extract features $f_1$ and $f_2$ from complete point clouds. Then we train the completion network (upper part) by jointly optimizing the feature matching loss $L_{feat}$, reconstruction loss $L_{recon}$ and adversarial loss $L_{GAN}$.}
  \vspace{-6mm}
  \label{overall_pipeline}
\end{figure*}

\section{Related work}

\subsection{Image Transformation}
Deep learning has shown a great success on image transformation from one domain to another domain, of which, conditional generative adversarial (cGAN) model~\cite{mirza2014conditional} is a popular approach. For example, Domain Transfer Network (DTN)~\cite{taigman2016unsupervised} and UNIT~\cite{liu2017unsupervised} work on translating the face to digit images, for both low and high resolutions. ~\cite{shrivastava2017learning} proposed a cGAN based method to map the rendered images to real images for gaze estimation. Cycle-consistency methods~\cite{zhu2017unpaired,kim2017learning} are proposed for the unsupervised image transformation and achieve impressive results. Apart from GAN-based methods, some works utilize Siamese networks~\cite{bromley1994signature,hu2018cvm,feng20192d3d} to calculate the triplet loss for image matching. 
However, existing approaches focus on synthesizing high-quality images, and are not suitable for fine-grained feature matching and transformation, which is our primary goal. Although several works~\cite{xian2019f,xian2018feature} show feature generation ability on few-shot tasks, they are conditioned on category labels or attributes, which is not robust to unseen classes.

\subsection{Deep Learning on 3D Data}
There is a large amount of impressive research on 3D learning with deep neural networks, and they can be categorized into three classes: voxels, meshes and point clouds. 
Many existing works~\cite{stutz2018learning,chang2015shapenet,wu20153d} conduct 3D analysis on voxels. However, 3D learning on voxels consumes a lot of memory, hence it is difficult to generate high resolution outputs. Even though some approaches~\cite{wang2017cnn,tatarchenko2017octree} propose to utilize octree structure to save memory, their results are not realistic enough.
Although 3D learning on meshes~\cite{vakalopoulou2018atlasnet,wang20193dn} show insightful results on complicated objects reconstruction, it is difficult to add new vertices to a mesh topology which is fixed in the beginning.
Point cloud has received growing attention in recent years and extensive works~\cite{qi2017pointnet,qi2017pointnet++,dgcnn} have been proposed to analyze point clouds. The pioneering work PointNet~\cite{qi2017pointnet} achieved great success on point cloud classification and segmentation. The key idea is to use a permutation invariant symmetric function to extract global features on the orderless points. In order to study local structures, PointNet++~\cite{qi2017pointnet++} propose a hierarchical architecture to learn geometric features in the local areas, and achieves superior results. 
However, the above methods are focused on the clean and complete input points, few methods are proposed for the incomplete and sparse point clouds. 
Several works~\cite{yu2018pu,yu2018ec,Yifan_2019_CVPR,li2019pugan} have proposed the point cloud upsampling methods to generate dense point clouds from sparse inputs. PU-Net~\cite{yu2018pu} adopts the PointNet++ as backbone to extract point features and upsamples the point size by consecutive convolutions. EC-Net~\cite{yu2018ec} generates more points on the edge area by calculating distance with edge labels. PU-GAN~\cite{li2019pugan} proposes an adversarial learning method to recover the dense points. 
Even though they are able to densify the sparse point clouds, they fail to complete the missing parts if the objects are incomplete.
Yuan et~al.~\cite{yuan2018pcn} proposes a point completion network PCN, which is a simple encoder-decoder architecture and is able to generate complete points given noise and incomplete inputs. TopNet~\cite{topnet2019} proposes a tree-structure decoder to synthesize the points in a hierarchical manner. MSN~\cite{liu2019morphing} proposes a coarse to fine strategy to reconstruct complete points. However, existing methods failed to generate fine-grained details of object shape. 
Some methods~\cite{mescheder2019occupancy,park2019deepsdf,michalkiewicz2019deep} propose to learn the reconstruction in a function space, but they only optimize the final reconstruction loss and ignore the constraint on the intermediate features, which are studied by our approach.

\section{Method}
We propose an encoder-decoder based network and adopt a two-stage training process. In the first stage, we learn the complete point feature embeddings $f_1\in R^{256}$ and $f_2\in R^{1024}$ by training a point cloud auto-encoder~\cite{yuan2018pcn}. 
In the second stage, we reload the decoder of the auto-encoder and optimize the whole network end-to-end.
To learn the shape prior from the complete objects, incomplete point features are extracted by the encoder sub-network and aligned with the complete points features by our feature alignment strategies. The aligned features are then fed into the decoder to reconstruct the dense and complete point clouds. We further apply the coarse-to-fine strategy to incorporate the shape prior from the partial inputs. Specifically, we concatenate the incomplete input and the coarse output in the fine stage such that the detailed information can be preserved. The overall pipeline is shown in Fig.~\ref{overall_pipeline}.

\subsection{Auto-encoder}
In order to learn the shape prior from the dataset, we train an auto-encoder network by optimizing the Chamfer Distance. 
The input is the ground truth point cloud $\textbf{Y}$ and the output is the reconstructed point cloud $\textbf{\^{Y}}$.
For point sets $\textbf{Y}$, an encoder is learnt to map $\textbf{Y}$ from the original point space to low-dimensional embedding spaces. 
A decoder inversely transforms the latent embedding back to the reconstructed point set $\textbf{\^{Y}}$. 
Once trained, the weights of the encoder and decoder are fixed. Two latent codes are obtained and they implicitly capture the manifold of the dense and complete point clouds.
We choose the PCN~\cite{yuan2018pcn} as our auto-encoder architecture.
Then we train the completion pipeline to map the partial point features to this latent space. 

\subsection{Encoder Sub-network}
Similar to PCN~\cite{yuan2018pcn}, we adopt the two-step point feature extraction method with max-pooling operation in the end of each step. We denote the two feature embeddings obtained by the max-pooling layers as $f_1\in R^{256}$ and $f_2\in R^{1024}$, respectively. 

\subsection{Feature Alignment Strategies}
\noindent\textbf{Feature Matching Loss}
Many image related works~\cite{tang2018ba, bloesch2018codeslam} demonstrate the effectiveness and superiority by considering feature metrics on intermediate outputs from different layers.
In view of this, we propose our feature matching strategy for point clouds. 
The latent representations of partial and corresponding complete point clouds are matched. For the latent loss, we experiment with the L2 and L1 distances and choose L2 because of the better results.
Hence, we directly calculate the L2 distance on two sets of features $f_{1}$, $f_2$ from $\textbf{X}$ and $\textbf{Y}$, as expressed by
\begin{align}
    \mathcal{L_{\text{feat}}} = \sum_{ij \in \mathcal{P}}\|f_{\textbf{X}_i} - f_{\textbf{Y}_j}\|_2,
\end{align}
where $\mathcal{P}$ denotes all pairs from the two feature sets $\{f_{\textbf{X}}, f_{\textbf{Y}}\}$, and $\textbf{X}$ and $\textbf{Y}$ represent the partial and complete point clouds, respectively.

\noindent\textbf{Adversarial Loss} 
We adopt the Maximum Mean Discrepancy Generative Adversarial Network (MMD-GAN)~\cite{li2017mmd,binkowski2018demystifying} to learn more accurate feature distributions.
The squared MMD distance between two distributions $P$ and $Q$ is calculated as
$M_k^2(P,Q)=\mathbb{E}_{a,a^{'} \sim P}[k(a,a^{'})]+\mathbb{E}_{b,b^{'} \sim Q}[k(b,b^{'})]-2\mathbb{E}_{a \sim P,b \sim Q}[k(a,b)]$,
in which $k(a,b)$ is a rational quadratic kernel~\cite{binkowski2018demystifying} calculating the similarity between two samples.
Therefore, we obtain the G loss function as $\mathcal{L}_{\text{GAN}}(G)=\mathbb{E}_{P_{G}} [k_D(b,b^{'})]-2\mathbb{E}_{P_R,P_G}[k_D(a,b)]+\mathbb{E}_{P_R} [k_D(a,a^{'})]$, and D loss as $\mathcal{L}_{\text{GAN}}(D)=-\mathcal{L}_{\text{GAN}}(G)+{\mathbb{E}}\big[(||\nabla_{\hat{x}}D(\hat{x})||_2-1)^2\big]$, in which $a$, $P_R$, $b$ and $P_G$ represents the real data, the real data distribution, the generated data and corresponding distribution, respectively.
$\hat{x}=\theta b + (1-\theta) a$, with $\theta \sim (0,1)$.
In our experiments, we compare the MMD-GAN approach with the widely used LS-GAN~\cite{mao2017least} in point clouds to show the advantage of MMD metric.

\begin{figure*}
\centering
  \includegraphics[width=0.95\linewidth]{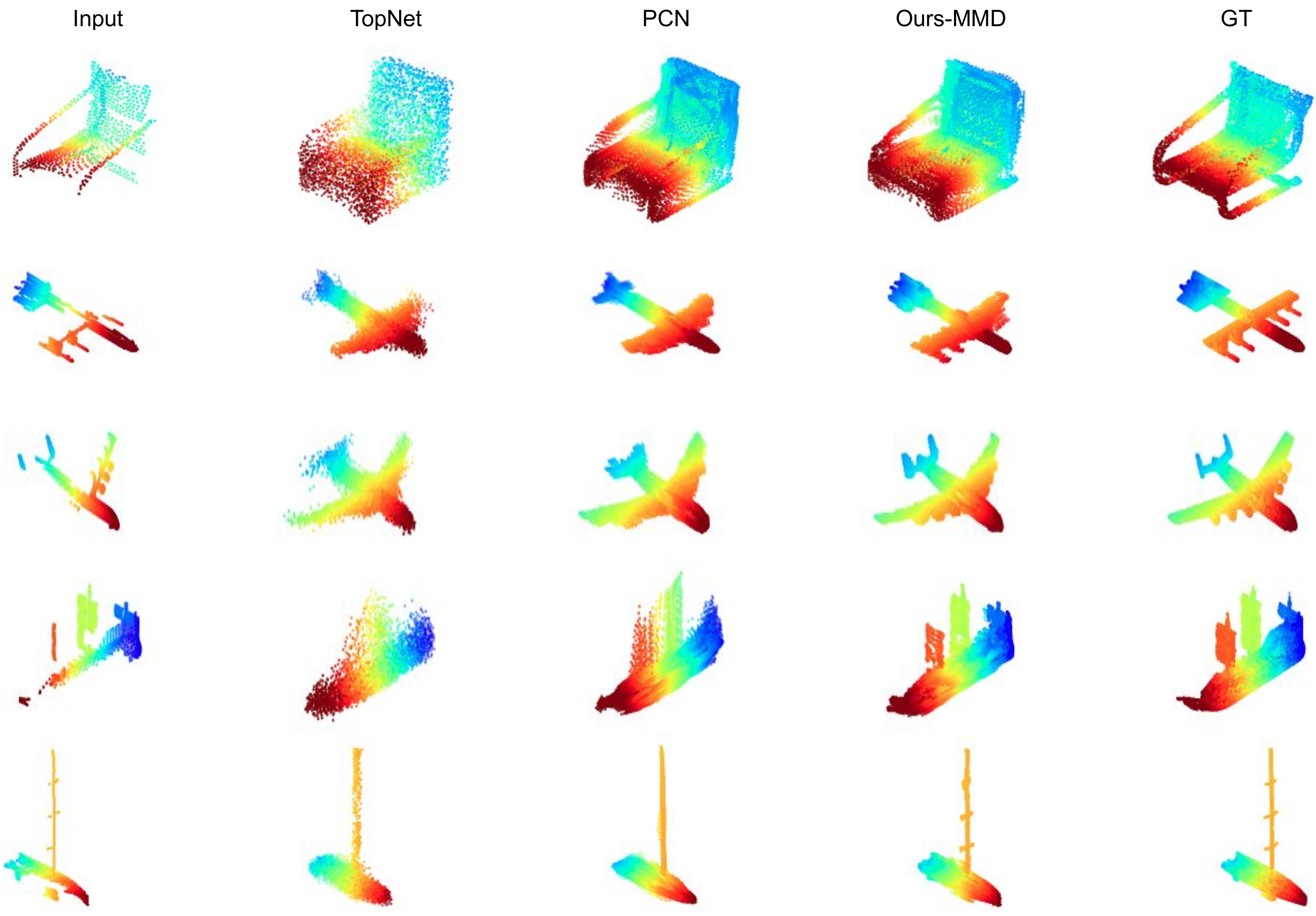}
  \caption{Qualitative results on the ShapeNet dataset.}
  \label{completion_qual}
\end{figure*}
\begin{table*}[!htbp]
\centering
\caption{Quantitative comparison for point cloud completion on seen categories of ShapeNet.}
\resizebox{0.90\textwidth}{!}{
\small
\centering
\begin{tabular}{l|c| c |c  |c  | c|  c|  c |c|c }
\hline
\multirow{2}{*}{Method} & \multicolumn{9}{c}{Mean Chamfer Distance per point ($10^{-3}$)} \\
\cline{2-10}
{}& Avg & Airplane & Cabinet & Car & Chair & Lamp & Sofa & Table & Vessel \\
\hline\hline
3D-EPN\cite{dai2017shape} &20.147 &13.161 &21.803 &20.306 &18.813 &25.746 &21.089 &21.716 &18.543  \\
PCN-FC\cite{yuan2018pcn} &9.799 &5.698 &11.023 &8.775 &10.969 &11.131 &11.756 &9.320 &9.720 \\
PCN\cite{yuan2018pcn} &9.636 &5.502 &10.625 &8.696 &10.998 &11.339 &11.676 &8.590 &9.665 \\
TopNet~\cite{topnet2019} &9.890 &6.235 &11.628 &9.833 &11.498 &9.366 &12.347 &9.362 &8.851 \\
Ours-L2 &8.574  &4.922  &\textbf{9.970}  &8.340  &9.519 &9.160  &10.707  &7.794 &8.177 \\
Ours-MMD & \textbf{8.496} &\textbf{4.773} &10.120 &\textbf{8.327} &\textbf{9.320} &\textbf{9.053} &\textbf{10.471} &\textbf{7.767} &\textbf{8.137}\\

\hline
\end{tabular}
}
\label{completion_quan_seen}
\end{table*}

\subsection{Decoder Sub-network}
We use the coarse to fine strategy to reconstruct the complete point clouds~\cite{yuan2018pcn}. In the first stage, coarse point clouds $\textbf{P}_{\textbf{coarse}}\in R^{\text{1024}\times \text{3}}$ are obtained by feeding the global feature $f_{2}$ into three fully-connected layers [1024,1024,3072] followed by a reshaping operation. 
The coarse point clouds represent the whole shape of an object, but some details are missing as shown in Fig.~\ref{overall_pipeline}.

In order to preserve the input object shape details and consider the long-range dependencies, we propose to concatenate the input partial points with the coarse results $\textbf{P}_{\textbf{coarse}}$ similar with~\cite{liu2019morphing,Wang_2020_CVPR}. 
Since the majority of objects are symmetric with respect to the \emph{x-y} plane, we adopt the mirroring operation~\cite{wang20193dn,kanazawa2018learning,Wang_2020_CVPR} on the partial input and then subsample the mirrored objects to obtain 512 points by the farthest point sampling (FPS)~\cite{qi2017pointnet++} algorithm. 
Thus, both the global and local object information are incorporated during the second stage of the decoder. 

In the second stage, we repeat the global feature $f_{2}$ by N times and tile $\textbf{P}_{\textbf{coarse}}^{'}$ by $\frac{\text{N}}{\text{1024}}$ times, then they are concatenated with a set of sampled grid coordinates to obtain a feature $f_{3}\in R^{\text{N}\times 1029}$.
Finally, we transform $f_{3}$ to the dense and complete output by MLPs with kernel size of [512,512,3]. We generate four resolutions outputs, i.e. N$\subseteq$ $\{\text{2048}, \text{4096}, \text{8192}, \text{16384}\}$.

\subsection{Optimization}
\noindent\textbf{Reconstruction Loss} 
There are two permutation invariant metrics proposed by \cite{fan2017point} to measure the distance between two unordered point sets: the Chamfer distance (CD) and the Earth Mover's distance (EMD).
We use CD to calculate the reconstruction loss because the maximum size of our dense output is 16384, which makes it impractical to use EMD. 
CD is calculated as, i.e., 
\begin{align}
    \begin{split}
    \mathcal{L}_{S_1,S_2}&=\frac{1}{|S_1|}\sum_{x\in S_1}  \min\limits_{y\in S_2}||x-y||_2,\\
    \mathcal{L}_{S_2,S_1}&=\frac{1}{|S_2|}\sum_{y\in S_2}  \min\limits_{x\in S_1}||x-y||_2,\\
    \text{CD}(S_1,S_2)&=\mathcal{L}_{S_1,S_2}+ \mathcal{L}_{S_2,S_1},\\
    \end{split}
\end{align}
in which $S_1$ and $S_2$ represent two point sets. 
Our reconstruction loss can be expressed as:
\begin{align}
    \mathcal{L}_{\text{recon}}=\text{CD}({\textbf{P}}_{\text{coarse}},\textbf{Y})+\lambda_\text{f} \text{CD}({\textbf{P}}_{\text{fine}},\textbf{Y}),
\end{align}
where ${\textbf{P}}_{\text{coarse}}$ and ${\textbf{P}}_{\text{fine}}$ represent the coarse output and finer output, respectively. $\lambda_\text{f}$ is the weight for the reconstruction loss of $\textbf{P}_{\text{fine}}$.

Our overall training loss includes three components: the reconstruction loss between the generated point cloud and the ground truth; the adversarial loss and the feature matching loss between the partial and complete point features, i.e., 
\begin{align}
    \mathcal{L}=\lambda_{\text{re}} \mathcal{L}_{\text{recon}}+ \lambda_{\text{gan}} \mathcal{L}_{\text{GAN}} + \lambda_{\text{fe}} \mathcal{L}_{\text{feat}},
\end{align}
where $\lambda_{\text{re}}$, $\lambda_{\text{gan}}$, and $\lambda_{\text{fe}}$ are the weights for three losses.

\begin{figure*}
\centering
  \includegraphics[width=1\linewidth]{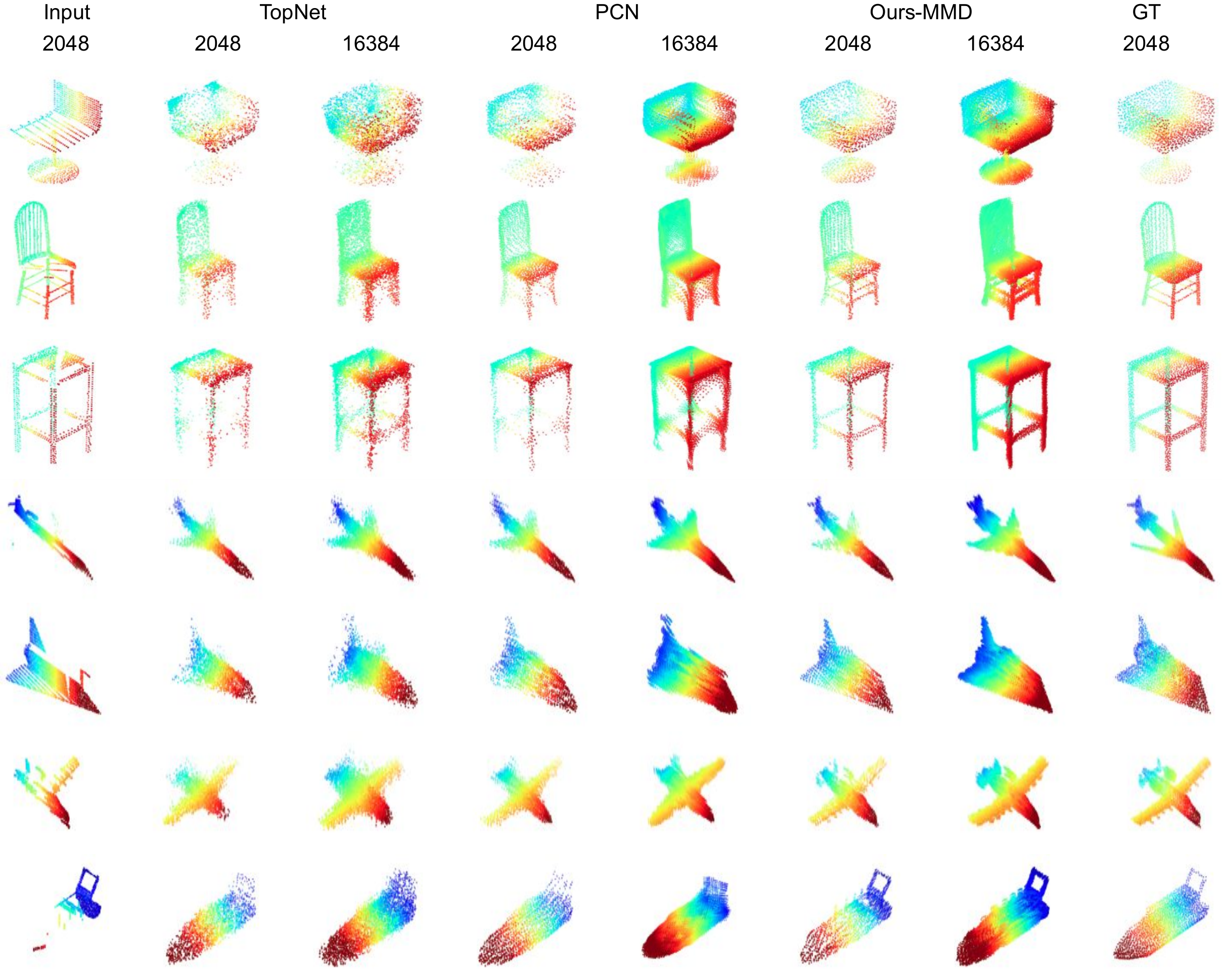}
  \caption{Shape completion results on 2048 and 16384 resolutions from the smaller dataset.}
  \label{qualitative_ours_new}
\end{figure*}
\begin{table*}[!htbp]
\centering
\vspace{-0.2cm}
\caption{Quantitative results for point cloud completion on novel categories of ShapeNet.}
\vspace{-0.2cm}
\resizebox{0.93\textwidth}{!}{
\small
\centering
\begin{tabular}{l|c c c  c   c|  c c c c c }
\hline
\multirow{2}{*}{Method} & \multicolumn{10}{c}{Mean Chamfer Distance per point ($10^{-3}$)} \\
\cline{2-11}
{}& Avg & Bus & Bed & Bookshelf & Bench & Avg & Guitar & Motorbike & Skateboard & Pistol \\
\hline\hline
3D-EPN\cite{dai2017shape} &41.5 &35.94 &47.85 &39.12 &43.07 &44.3 &47.35 &40.67 &47.84 &41.36  \\
PCN\cite{yuan2018pcn} &14.2 &9.46 &21.63 &14.79 &11.02 &12.9 &10.40 &14.75 &12.04 &14.23 \\
PCN-Folding\cite{yuan2018pcn} &13.8 &10.58 &19.08 &14.88 &10.55 &12.4 &9.06 &15.56 &11.91 &13.13 \\
TopNet~\cite{topnet2019} &14.1 &10.61 &\textbf{18.50} &14.93 &12.23 &\textbf{10.9} &\textbf{7.62} &13.50 &\textbf{8.92} &13.51 \\
Ours-L2 &13.4 &9.28 &20.38 &13.48 &\textbf{10.32} &11.3 &10.45 &\textbf{12.01} &10.77 &12.30 \\
Ours-MMD &\textbf{13.4} &\textbf{9.22} &20.64 &\textbf{13.39} &10.48 &11.0 &8.54 &12.56 &12.59 &\textbf{12.18} \\

\hline
\end{tabular}
}
\vspace{-0.4cm}
\label{completion_quan_unseen}
\end{table*}

\section{Experiments}

\subsection{Evaluation Metrics} We compare our method with existing state-of-the-art approaches by evaluating the Chamfer Distance (CD) on the synthetic dataset following~\cite{yuan2018pcn,topnet2019}.
PCN and TopNet calculate CD differently, we denote as CD-P and CD-T. 
We follow their evaluation methods on different settings for fair comparisons. We use CD-P in the PCN dataset and use CD-T for the remaining experiments.
Apart from CD, fidelity error (FD) is calculated to compare the results on the real KITTI dataset~\cite{geiger2013vision} following PCN~\cite{yuan2018pcn}.  

\begin{table}[!htbp]
\centering
\small
\caption{Quantitative comparison on the smaller training dataset with CD ($10^{-4}$). 
}
\vspace{-0.2cm}
\begin{tabular}{l|c| c  |c|c}
\hline
\multirow{2}{*}{Methods} & \multicolumn{4}{c}{Resolution} \\
\cline{2-5}
{}& 2048 &4096 &8192 &16384  \\
\hline
PCN~\cite{yuan2018pcn}  & 9.02 &7.71 &6.90 &6.17 \\
TopNet~\cite{topnet2019}  &9.88  &8.52 &7.56  &6.60 \\
Ours-L2  &7.42  &\textbf{6.40}  &5.84 &\textbf{5.13} \\
Ours-MMD  &\textbf{7.39}  &6.54  &\textbf{5.80} &5.20 \\
\hline
\end{tabular}
\vspace{-0.5cm}
\label{quantitative_ours}
\end{table}
\begin{table*}[!htbp]
\small
\centering
\caption{CD ($10^{-4}$) comparisons among different methods. The lower the better.} 
\resizebox{0.90\textwidth}{!}{
\begin{tabular}{l|c| c| c| c| c|c| c| c} 
\hline
Method &  PCN~\cite{yuan2018pcn} &  PCN~\cite{yuan2018pcn} + L2 &PCN~\cite{yuan2018pcn} + LS &PCN~\cite{yuan2018pcn} + MMD & Ours-BS & Ours-L2 & Ours-LS & Ours-MMD \\  
\hline
CD  &9.02 & 8.38 &8.54 &8.46 &7.85 &7.42 &7.59 &\textbf{7.39}\\
\hline
\end{tabular}
}
\vspace{-0.3cm}
\label{ablation}
\end{table*}
\subsection{Datasets} 
\vspace{-0.4cm}
Following PCN~\cite{yuan2018pcn}, both the synthetic dataset ShapeNet~\cite{chang2015shapenet} and the real-world dataset KITTI~\cite{geiger2013vision} are used to validate the effectiveness of our method. For ShapeNet, 30974 objects from eight categories are selected: airplane, cabinet, car, chair, lamp, sofa, table and vessel. Complete object has 16384 points and incomplete object has 2048 points.   
To further verify the generality of our network, we also test on eight another unseen categories of ShapeNet: bed, bench, bookshelf, bus, guitar, motorbike, pistol and skateboard. 
We test our method on a sequence of Velodyne scans from the KITTI dataset \cite{geiger2013vision} for the car object completion. 
Apart from the diverse training data of PCN, we test on a smaller training data~\cite{Wang_2020_CVPR}, in which the data is one eighth of data from PCN,
since one virtual view instead of eight views are rendered as the partial scans.
This is to evaluate the generality and robustness of our method. We generate four resolution results following TopNet~\cite{topnet2019}.

\subsection{Implementation Details}
Our models are trained by the Adam~\cite{kingma2014adam} optimizer. The initial learning rate is 0.0001 and is decayed by 0.7 every 20 epochs. The batch size is 32. The loss weights for $\lambda_{\text{re}}$, $\lambda_{\text{gan}}$  and $\lambda_{\text{fe}}$ are 200, 1 and 1000 respectively. $\lambda_\text{f}$ increases from 0.01 to 1 in the first 50000 iterations. We train one single network for all categories.

\subsection{Results}
We compare our method against existing state-of-the-art point clouds based methods 3D-EPN~\cite{dai2017shape}, PCN~\cite{yuan2018pcn} and TopNet\cite{topnet2019}.
We denote our method without feature alignment strategies as baseline method ``Ours-BS", with L2 loss as ``Ours-L2", with LS-GAN adversarial loss as ``Ours-LS" and with MMD-GAN adversarial loss as ``Ours-MMD".

\noindent\textbf{PCN Dataset}\label{exp_pcn}
Quantitative and qualitative results on PCN dataset are shown in Tab.~\ref{completion_quan_seen}, Tab.~\ref{completion_quan_unseen} and Fig.~\ref{completion_qual}. The object categories in the testing data are the same with that in the training data in Tab.~\ref{completion_quan_seen}, while all testing data in Tab.~\ref{completion_quan_unseen} are unseen. 
Tab.~\ref{completion_quan_seen} and~\ref{completion_quan_unseen} show that our method outperforms existing approaches. For the seen categories, we obtain $11.8\%$ and $14.1\%$ improvements on the average value compared to PCN and TopNet, respectively. The qualitative results in Fig.~\ref{completion_qual} show that our method is able to generate finer object details compared to other approaches, e.g. thin legs of chair and airplane wings.  
Our method also shows good performance on the eight unseen categories as shown in Tab.~\ref{completion_quan_unseen}, which verifies the robustness and generality of our method. 

\noindent\textbf{Smaller Dataset}
We show qualitative and quantitative results on the smaller training data in Fig.~\ref{qualitative_ours_new} and Tab.~\ref{quantitative_ours}, respectively. Our method achieves the best performance on all four resolutions compared to
existing approaches. Moreover, according to Tab.~\ref{quantitative_ours}, the relative improvements of our method compared to PCN are $18.1\%$, $15.2\%$, $15.9\%$ and $15.7\%$ for all resolutions on the smaller training data. 
Similar trends are observed in the comparison of our method to TopNet.
Fig.~\ref{qualitative_ours_new} shows that our method is not only able to preserve the details in the partial inputs, but also can generate the missing parts with finer details for different resolution results. For example, we show better results on the legs of chairs (Row 2 and 3); wings and engines of an aeroplane (Row 6).

\noindent\textbf{Ablation Study}
We explore the effects of feature alignment strategies in this section. The training of ablation experiments are conducted on the smaller data with 2048 resolutions. The results are shown in Tab.~\ref{ablation}. 
The results indicate that the feature alignment strategies are beneficial for both PCN~\cite{yuan2018pcn} and our baseline method. 
Adopting the L2 loss alone decreases the CD errors. 
Best results are obtained by further adding the MMD adversarial optimization.

\begin{figure}
\centering
  \includegraphics[width=1\linewidth]{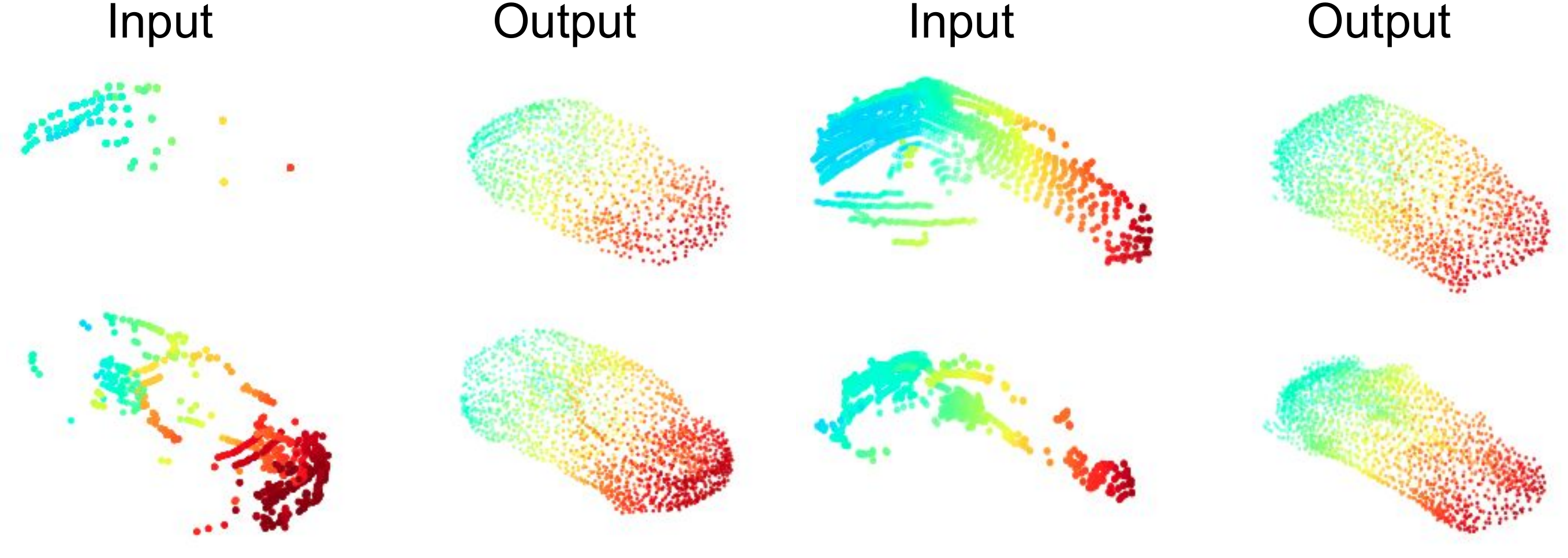}
  \vspace{-5mm}
  \caption{Qualitative results on the KITTI dataset.}
  \vspace{-5mm}
  \label{kitti_cars}
\end{figure}

\noindent\textbf{KITTI Dataset}
We also test our model on real scanned points to validate the generalization ability.  Following PCN~\cite{yuan2018pcn}, we make use of the fidelity error as the evaluation metric for different methods.
Fidelity error calculates the average distance from each point in the input to its nearest neighbour in the output.
Note that we directly use the models trained on the smaller dataset from all categories for testing, since there is no ground truth for the KITTI dataset.
All the models are trained for the resolution of 2048 points.
Quantitative comparison in Tab.~\ref{completion_kitti} shows that our feature optimization strategy outperform other approaches and can generate more accurate contextual information for cars.
This indicates that our strategies are more robust to noise and generalizable to real-world datasets.
Qualitative results in Fig.~\ref{kitti_cars} indicate that we can generate reasonable shapes even with few points in a partial scan.
\begin{table}[!htbp]
\centering
\small
\caption{Fidelity error (FD) comparison for point cloud completion on the KITTI dataset.}
\resizebox{0.5\textwidth}{!}{
\begin{tabular}{l|c| c| c | c | c }
\hline
Method &  PCN \cite{yuan2018pcn} & TopNet \cite{topnet2019} & Ours-L2 & Ours-LS & Ours-MMD\\
\hline
FD &0.0525 &0.0455 &\textbf{0.0339} &0.0661 &0.0340 \\
\hline
\end{tabular}
}
\vspace{-5mm}
\label{completion_kitti}
\end{table}

\section{CONCLUSIONS}
In this paper, we propose a point completion network to learn the shape priors from both the complete and partial points.
We propose feature alignment strategies to learn the complicated point distributions by a feature matching loss and an adversarial optimization with the help of MMD-GAN.
Various experiments show that our method achieves superior performance compared to state-of-the-art approaches on the point completion task.





\section*{ACKNOWLEDGMENT}
This research was partially supported by the Singapore MOE Tier 1 grant R-252-000-A65-114, the National University of Singapore Scholarship Funds and the National Research Foundation, Prime Minister’s Office, Singapore, under its CREATE programme, Singapore-MIT Alliance for Research and Technology (SMART) Future Urban Mobility (FM) IRG. 





                                  
\clearpage

\bibliographystyle{IEEEtran}
\balance
\bibliography{IEEEabrv,IEEEexample}

\end{document}